\pdfoutput=1

\documentclass[11pt]{article}

\usepackage[]{EMNLP2022}

\usepackage{times}
\usepackage{latexsym}
\usepackage{makecell}

\usepackage[T1]{fontenc}

\usepackage[utf8]{inputenc}

\usepackage{microtype}

\usepackage{inconsolata}

\usepackage{verbatim}
\usepackage{geometry}
\usepackage{pdflscape}
\usepackage{dirtytalk}
\usepackage[normalem]{ulem}
\usepackage{booktabs}
\usepackage{tikz}
\usepackage{amssymb}
\usepackage{pifont}
\usepackage{algpseudocode}
\usepackage[normalem]{ulem}
\useunder{\uline}{\ul}{}

\usepackage{todonotes}

\title{MT-GenEval: A Counterfactual and Contextual Dataset for Evaluating Gender Accuracy in Machine Translation}

\author{Anna Currey,
        Maria N\u{a}dejde,
        Raghavendra Pappagari,
        Mia Mayer,\\
        \textbf{Stanislas Lauly,}
        \textbf{Xing Niu,}
        \textbf{Benjamin Hsu,}
        \textbf{Georgiana Dinu}\\
        AWS AI Labs\\
        \texttt{ancurrey@amazon.com}}

\begin{document}
\maketitle
\begin{abstract}

As generic machine translation (MT) quality has improved, the need for targeted benchmarks that explore fine-grained aspects of quality has increased~\citep{freitag-etal-2021-experts,isabelle-etal-2017-challenge}. 
In particular, gender accuracy in translation \citep{choubey-etal-2021-gfst,saunders-byrne-2020-reducing} can have implications in terms of output fluency, translation accuracy, and ethics. 
In this paper, we introduce \textit{MT-GenEval}, a benchmark for evaluating gender accuracy in translation from English into eight widely-spoken languages. 
MT-GenEval complements existing benchmarks by providing realistic, gender-balanced, counterfactual data in eight language pairs where the gender of individuals is unambiguous in the input segment, including multi-sentence segments requiring inter-sentential gender agreement. 
Our data and code is publicly available under a CC BY SA 3.0 license.\footnote{\url{https://github.com/amazon-research/machine-translation-gender-eval}} 
\end{abstract}

\section{Introduction}
\label{sec:intro}

Although neural machine translation (NMT) has made great strides in quality~\citep{hassan2018achieving,wu2016google}, evaluations on generic test sets may not tell the whole story. 
Indeed, NMT models are known to make systematic errors in areas like robustness to input perturbations~\citep{niu-etal-2020-evaluating}, disambiguating pronouns in context~\citep{muller-etal-2018-large}, 
and translating unambiguous human gender~\citep{stanovsky-etal-2019-evaluating}. 
In particular, gender-related issues in translation can lead to translations that are inaccurate, ungrammatical, or biased. 

Accordingly, there has been increasing interest in improving machine translation for gendered entities~\citep{bentivogli-etal-2020-gender,choubey-etal-2021-gfst,saunders-byrne-2020-reducing,savoldi-etal-2021-gender,stanovsky-etal-2019-evaluating}. Adequate evaluation benchmarks play an important role in supporting this line of research and improving understanding of how models perform on the task of gender translation accuracy. Existing benchmarks have limited diversity in terms of gender phenomena (e.g., focusing on professions), sentence structure (e.g., using templates to construct sentences), or language coverage (see section~\ref{sec:compare} for more information), making it difficult to gauge how systems perform in terms of both gender and quality simultaneously.

\begin{table*}
    \centering\small
    \begin{tabular}{l|l}
         &  \textbf{Segment Pair}\\\hline
        1 & After some wrangling Blacket accepted £50 in full settlement of the fees due to \underline{him}.\\
        & Nach einigem Hin und Her akzeptierte Blacket 50 Pfund als vollen Ausgleich für die \underline{ihm} zustehenden Gebühren.\\[2pt]
        2 & Having served \underline{his} apprenticeship Crookall became a \textit{master painter} trading at Duke Street, Douglas.\\
        & Tras servir como aprendiz, Crookall se convirtió en \underline{maestro pintor} en la calle Duke de Douglas.\\[2pt]
        3 & Serrano agreed to the restaurant contract as long as \underline{he} could have a tapas restaurant. Serrano \textit{traveled} to Spain [\dots]\\
        & Serrano s'est \underline{rendu} en Espagne [\dots]\\
    \end{tabular}
    \caption{Examples of segment pairs that are included in MT-GenEval. Explicitly gendered words are \underline{underlined}, while ungendered words whose translation is explicitly gendered are in \textit{italics}.}
    \label{tab:intro-examples}
\end{table*}

To this end, this paper releases a new \textbf{M}achine \textbf{T}ranslation \textbf{Gen}der \textbf{Eval}uation benchmark: \textbf{MT-GenEval}. 
We include development data (for model improvements) as well as test data and corresponding metrics (for comprehensive evaluation). 
MT-GenEval is realistic and diverse, includes a wide variety of contexts for gender disambiguation, and is fully balanced by including human-created gender counterfactuals. 
In its first release, the MT-GenEval benchmark covers translation from English into eight target languages, for two genders.

MT-GenEval focuses on the task of \textbf{gender accuracy in translation}. 
We define gender accuracy in translation as the extent to which a machine translation output accurately reflects the gender of the humans mentioned in the input, restricted to cases where the gender is explicitly and linguistically disambiguated in the context of the input. Thus, in our benchmark we do not consider the grammatical gender on inanimate objects, or cases where the input gender is ambiguous within the given level of context.\footnote{In this latter case, multiple translations are valid, so this falls under the purview of gender \textit{customization} tasks~\citep{habash-etal-2019-automatic,nadejde-etal-2022-coca-mt,saunders-etal-2020-neural,vanmassenhove-etal-2018-getting}.} 

Table~\ref{tab:intro-examples} gives examples of the data in the MT-GenEval dataset. In all cases, the source segment contains a reference to a human and that human's gender is unambiguous based on the linguistic context, be that context intra-sentential (rows 1 and 2), or inter-sentential (row 3). 
Row 2 shows a case where although a given entity (\textit{Crookall}) is gendered in both the source and the target, the gendered words themselves are different (\textit{his} is marked for gender in English but absent in Spanish, while \textit{master painter} is marked for gender in Spanish but not English). 
This is in large part what makes gender accuracy in translation non-trivial, especially in real, diverse, and long segments.

We provide a detailed description of the dataset in section~\ref{sec:gender8desc} and of the associated automatic evaluation metrics in section~\ref{sec:metrics}. Section~\ref{sec:compare} gives an overview of existing gender accuracy benchmarks for machine translation; we hope this overview will  
enable researchers to assess which benchmarks are appropriate for their use cases. 
Finally, section~\ref{sec:benchmark} evaluates commercial systems as well as models trained on public data on MT-GenEval. We find that: (1) systems trained with contextual and gender-filtered data show improvements in both inter- and intra-sentential gender accuracy as measured by MT-GenEval; and (2) generic (unrelated to gender) translation quality is correlated with gender, exhibiting a new facet of gender in machine translation that is understudied in prior work.

\begin{table*}[t!]
\resizebox{\textwidth}{!}{
\begin{tabular}{l|p{5.3cm}|p{5.3cm}|p{5.3cm}|p{5.3cm}}
 & \textbf{Feminine Source} &  \textbf{Feminine Reference} & \textbf{Masculine Source} & \textbf{Masculine Reference} \\
\hline
1  & \underline{Her} family moved to the midwest where \underline{she} was educated and permanently scarred by dour \underline{nuns}.
& Sa famille a déménagé dans le Midwest où \underline{elle} a été \underline{éduquée} et irrémédiablement \underline{traumatisée} par des \underline{religieuses} austères.
& \underline{His} family moved to the midwest where \underline{he} was educated and permanently scarred by dour \underline{monks}. 
& Sa famille a déménagé dans le Midwest où \underline{il} a été \underline{éduqué} et irrémédiablement \underline{traumatisé} par des \underline{moines} austères.\\
\hline
 2 & Many of \underline{her} short stories have been broadcast on BBC Radio 4. 
 & Muchos de sus relatos cortos se han emitido en BBC Radio 4. 
 & Many of \underline{his} short stories have been broadcast on BBC Radio 4.
 & Muchos de sus relatos cortos se han emitido en BBC Radio 4.\\
\end{tabular}}
\caption{Counterfactual examples from MT-GenEval. Each source segment refers to individuals of a single gender (in this case, female or male), and counterfactual segments change all unambiguous gender references. 
Unambiguous gendered words are \underline{underlined}. In some cases, the reference translation might not be gendered (row 2).}
\label{tab:examples}
\end{table*}

\begin{table*}
\small
\resizebox{\textwidth}{!}{
\begin{tabular}{l|p{5cm}|p{5cm}|p{5cm}}
 & \textbf{Context and Source} & \textbf{Correct Reference} & \textbf{Contrastive Reference} \\
\hline
1 & Paul intervenes and overpowers \underline{him}, but \underline{he} wriggles free. <sep> \textit{The librarian} is then \textit{run over} by a car in front of the library and apparently killed. & \underline{El bibliotecario} es luego \underline{atropellado} por un auto enfrente de la biblioteca y al parecer murió. & \underline{La bibliotecaria} es luego \underline{atropellada} por un auto enfrente de la biblioteca y al parecer murió.\\
\hline
2 & After the war, \underline{she} continued \underline{her} career at the Boruprokat factory. <sep> Hasanova was \textit{the} chief \textit{brigadier} in 1970 and led four brigades at that factory. & Hasanova era \underline{la brigadiera} capo nel 1970 e guidò quattro brigate nella stessa fabbrica. & Hasanova era \underline{il brigadiere} capo nel 1970 e guidò quattro brigate nella stessa fabbrica. \\
\end{tabular}
}
\caption{Contextual examples from MT-GenEval. Note that unlike counterfactual examples, reference examples are \textit{contrastive}. Contrastive references are available for the main sentence (which comes after \textit{<sep>}), but not for the context. Unambiguous gendered words are \underline{underlined}, and their ambiguous translations are in \textit{italics}.}
\label{tab:examples-contextual}
\end{table*}

\section{MT-GenEval: Gender Translation Accuracy in 8 Language Pairs}
\label{sec:gender8desc}

For the initial release, MT-GenEval covers translations in two genders (female and male)\footnote{We recognize that the coverage of only two genders is a limitation of our work. As such, we plan to expand to additional genders in the future.} from English (EN) into eight diverse and widely-spoken target languages: Arabic (AR), French (FR), German (DE), Hindi (HI), Italian (IT), Portuguese (PT), Russian (RU), and Spanish (ES). 
The source language has limited morphological gender, with gender expressed only on some pronouns and nouns. 
By contrast, the target languages have extensive grammatical gender and may express gender through morphological markings on a variety of parts of speech including verbs and adjectives, as well as on inanimate objects. In the target languages, human gender often, but not always, lines up with grammatical gender. 
In order to facilitate evaluation and training of gender-accurate machine translation systems, we release two test subsets (\textit{counterfactual}, Table~\ref{tab:examples}; and \textit{contextual}, Table~\ref{tab:examples-contextual}) as well as a counterfactual development set. Each subset is described in more detail below.

\subsection{Counterfactual Subset}
\label{sec:gender8desc-create}

\paragraph{Data sourcing}
In developing MT-GenEval, our goal was to create a realistic, gender-balanced dataset that naturally incorporates a diverse range of gender phenomena. 
To this end, we extracted English source sentences from Wikipedia\footnote{\url{https://www.wikipedia.org/}} as the basis for our dataset. 
We automatically pre-selected relevant sentences using EN gender-referring words based on the list provided by~\citet{zhao-etal-2018-gender}. 
In addition to the sentence containing the relevant gendered word(s), we included the two prior sentences in the pre-selection, so as to increase the diversity of gendered words beyond the list used.
In a second stage, we asked annotators to manually review these initial candidate segments to ensure that they contain (1) at least one reference to an unambiguously gendered human, (2) no references to individuals of a different gender, and (3) no first names (to avoid confounds where models associate a first name with a gender). 
Items (2) and (3) are necessary to enable the creation of counterfactual source segments. 

\paragraph{Gender balance through counterfactuals}
In order to ensure that our dataset was fully balanced between female and male genders, as well as to eliminate correlations between gender and content, 
we asked annotators to manually create \textit{counterfactual} versions of each source segment. 
Since each source segment refers to individuals of a single gender (in our case, female or male), we were able to create counterfactual version by changing all unambiguous references to that gender (e.g., female) to equivalent unambiguous references to another gender (e.g., male). 
See Table~\ref{tab:examples} for an example of original and counterfactual sentences. 

\paragraph{Reference translations}
We asked professional translators to create translations for the original segments from scratch, and to use post-editing for the corresponding counterfactual segments, where annotators had access to both the counterfactual and the original source segment during post-editing. 
This had the effect of eliminating spurious differences in the original and counterfactual translations that were unrelated to gender. 
Translators were encouraged \textit{not} to introduce gender marking in the translation when such differences would not be natural. Thus, for HI, IT, and ES, several of the gendered inputs have gender-neutral translations. 

All annotation was done by professional linguists/translators who are native speakers of the relevant language (English in the case of sourcing and counterfactual creation; target languages in the case of translations). 
Additionally, annotations were reviewed by professional quality assurance teams to ensure that the data was high-quality. 
For each step in the process, half the data was created by a female annotator and half by a male annotator. 
We have released the full text of the annotation instructions on GitHub.\footnote{\url{https://github.com/amazon-research/machine-translation-gender-eval}}

\paragraph{Development and test data} 
The counterfactual \textbf{test} set consists of 600 segments (balanced by gender), all of which have gender-specific sources \textit{and} references. 
Each segment in the test set also has its counterfactual in the test set, which facilitates automatic evaluation (see section~\ref{sec:metrics}). 
We additionally release \textbf{development} data, which consists of 2400 sentence-level segments. 
Unlike the test data, we do not enforce that all reference translations in the development set be gendered. 
As such, 84.7\% of references are gendered for EN-HI, 89.0\% for EN-IT, and 89.2\% for EN-ES (for the remaining five language pairs, all references are gendered). 

\subsection{Contextual Subset}
\label{sec:gender8desc-subsets}

\paragraph{Data sourcing} In developing the {contextual} subset of MT-GenEval, our goal was to create a gender-balanced dataset for evaluating gender accuracy and bias in \textit{contextual} MT models. 
First, using word lists, we automatically pre-selected sentences from Wikipedia that contained at least one mention of a profession and no gendered words. The selected professions fall into one of three categories: stereotypical female, stereotypical male and neutral \citep{troles-schmid-2021-extending}. Additionally, the professions were selected to lack gender marking in English, while potentially requiring gender inflection and agreement in the target languages. To remove any further gender cues, sentences containing commonly used first names were excluded.\footnote{We used publicly available US and UK census data from  \url{https://github.com/OpenGenderTracking/globalnamedata}.} Annotators were subsequently asked to manually review the remaining segments to ensure that (1) they were indeed gender-ambiguous on the segment level and (2) they contained mentions of exactly one individual.

\paragraph{Context}
For each of the selected gender-ambiguous sentences, we extracted the two preceding context sentences. We took a semi-automatic approach to verifying whether the context sentences disambiguated the gender of the selected sentence, first checking for the presence of gendered words and subsequently asking annotators to mark which context sentence disambiguated the selected sentence (none, one of them, or both).
This yielded a set of gender-ambiguous sentences referring to a single individual along with at least one preceding context sentence that linguistically disambiguated the gender of that individual. We give examples of selected source sentences in Table~\ref{tab:examples-contextual}, where we use the \textit{<sep>} token to delimit the context and main sentence.

\paragraph{Gender balance}
To ensure the dataset was balanced, we selected an equal number of source examples for both female and male genders, as well as for each profession category.  As a result, the dataset contains both stereotypical and anti-stereotypical examples and covers six sub-categories in total, as shown in Table~\ref{tab:context-subsets}.

\begin{table}
    \centering
    \begin{tabular}{l|c|c}
     & \multicolumn{2}{l}{\textbf{Person Gender}} \\ \hline
    \textbf{Profession} &	Female & Male \\ \hline
Female & 150 & 150 \\ 
Male & 	150	& 150 \\
Neutral & 250 & 250 \\ \hline
\textbf{Total} & 550 & 550 \\ 
    \end{tabular}
    \caption{Number of source sentences in each of the six sub-categories of the contextual subset of MT-GenEval.}
    \label{tab:context-subsets}
\end{table}

\paragraph{Reference translations}
In addition to the original source segments, we release two contrastive reference translations for each main sentence (references for context sentences are not included in the dataset). One of the reference translations correctly translates the gender of the individual, while the contrastive reference changes the gender (e.g., female to male). For example, as shown in Table~\ref{tab:examples-contextual}, the ambiguous noun phrase ``the librarian'' is translated as ``el bibliotecario'' in the correct reference and as ``la bibliotecaria'' in the contrastive reference. We asked translators not to introduce unnecessary gender marking (similar guidelines as for the {counterfactual} subset) and we excluded from this subset examples where the contrastive translations were identical (no gender marking).

\section{Automatic Metrics for MT-GenEval}
\label{sec:metrics}

\subsection{Gender Accuracy}
\label{sec:metrics-sentences}

All segments in our test set include both correct and contrastive/counterfactual references. 
To automatically evaluate gender accuracy in translation on this set, we propose a straightforward accuracy metric based on the fact that, by design, the correct and contrastive references differ only in gender-specific words. 

We define accuracy of gender in translation on our test set as follows.\footnote{We considered other approaches to defining gender translation accuracy, e.g., based on model score on the correct vs.\ contrastive references. However, we found that these had lower agreements with human scores in initial experiments.}
Let $w_{hyp}$, $w_{ref}$, and $w_{con}$ denote the set of words in the hypothesis, reference, and contrastive reference, respectively. 
First, we obtain the set of words in the contrastive reference that are not in the correct reference:
\begin{equation}
    unique_{con} = w_{con} \setminus w_{ref}
\end{equation}
This removes from consideration all the words that are unrelated to the gender of the individual(s) in the source, as the correct reference and contrastive reference do not have any non-gender-related differences. 
We consider a segment \textit{incorrect} if: 
\begin{equation}
    unique_{con} \cap w_{hyp} \neq \emptyset
\end{equation}
i.e., if the hypothesis contains words specific to the contrastive (incorrect) gender. 

To evaluate our metric, we ran human evaluations of gender accuracy on a subset of the contextual set. We selected a stratified sample of 600 source segments, translated each with three commercial systems, and asked two professional translators to mark the gender correctness of the system outputs. Further details on the evaluation task are provided in section~\ref{context-benchmarking}, including inter-annotator agreement scores. 
Table~\ref{tab:metric-eval} shows average F-score of the automatic accuracy metric with respect to human annotations in this evaluation. 
The automatic metric matches humans reasonably well across all language pairs, with F-scores consistently at 0.80 or higher. 

\begin{table}[htb]
    \centering
    \begin{tabular}{l|c}
     & \textbf{F-score}\\\hline
    EN$\rightarrow$AR & 0.84\\
    EN$\rightarrow$DE & 0.82\\
    EN$\rightarrow$ES & 0.89\\
    EN$\rightarrow$FR & 0.86\\
    EN$\rightarrow$HI & 0.80\\
    EN$\rightarrow$IT & 0.83\\
    EN$\rightarrow$PT & 0.86\\
    EN$\rightarrow$RU & 0.84
    \end{tabular}
    \caption{F-score between automatic accuracy metric and human accuracy labels on the contextual set.}
    \label{tab:metric-eval}
\end{table}

Gender accuracy on the counterfactual subset is defined similarly.\footnote{Initial human evaluations on this set were unreliable, with very low inter-annotator agreements. As such, we leave evaluation of the metric on the counterfactual subset for future work, as this is a relatively difficult task for which annotators need more training.} 
However, on the counterfactual subset we have pairs of counterfactual source segments as well as counterfactual references. 
We consider a segment pair as correct only if \textit{both} the original and the counterfactual segment are marked as correct. 
This is to reward models for cases where they are actually predicting correct gender based on the input, rather than randomly guessing. 

\subsection{Gender Quality Gap}
\label{sec:metrics-quality}

While the gender accuracy metric introduced in section~\ref{sec:metrics-sentences} evaluates gender translation at the lexical level, 
\textit{generic translation quality} may also vary across the inputs and may be correlated with gender. For this reason, we complement the accuracy metric with a \textbf{gender quality gap} metric, $\mathbf{\Delta_{qual}}$. 
This allows us to measure representational bias~\citep{blodgett-etal-2020-language}, expressed as lower quality for one of the two genders considered, on MT-GenEval.

We evaluate $\mathrm{\Delta_{qual}}$ on the counterfactual subset, where we can abstract away non-gender-related content differences (since for a given sentence, its semantically equivalent gender-counterfactual is always in this test set). We define $\mathrm{\Delta_{qual}}$ as:
\begin{equation}
    \mathrm{\Delta_{qual} = BLEU_{male} - BLEU_{female}}
\end{equation}
where $\mathrm{BLEU_{gender}}$ is the BLEU score of the $gender$ subset of the counterfactual test set.

\begin{table*}[t!]
{
    \centering\small
    \begin{tabular}{p{2.6cm}|p{2.2cm}ccp{3.4cm}cc}
    \textbf{Benchmark} & \textbf{Languages} & \textbf{Size} & \textbf{Data Type} & \textbf{Advantages} & \textbf{References?} & \textbf{Metric?}\\ \hline  
    
    MT-GenEval (\textbf{ours}) & AR, DE, EN, ES, FR, HI, IT, PT, RU &  4,008   & natural & counterfactual sources and references; inter-sentential context; gender quality gap & \checkmark & \checkmark\\[2pt]
    
    GeBioCorpus \cite{costa-jussa-etal-2020-gebiotoolkit} & CA, EN, ES & 2,000 & natural & extraction pipeline & \checkmark & $\times$\\[2pt]
    
    MuST-SHE \cite{bentivogli-etal-2020-gender} & EN, ES, FR, IT & 1,095 & natural & counterfactual references; aligned audio & \checkmark & \checkmark \\[2pt]
    
    SimpleGEN \cite{renduchintala-etal-2021-gender} &  DE, EN, ES & 1,332 & synthetic & stereotype annotations & $\times$ & \checkmark\\[2pt]
    
    Translated Wikipedia Bios\footnote{\url{https://ai.googleblog.com/2021/06/a-dataset-for-studying-gender-bias-in.html}} & DE, EN, ES & 138 & natural & inter-sentential context & \checkmark & $\times$\\[2pt]
    
    WinoMT \cite{stanovsky-etal-2019-evaluating} & AR, CS, DE, EN, ES, FR, HE, IT, PL, RU, UK & 3,888 & synthetic & stereotype annotations; easily extensible to new target languages & $\times$ & \checkmark
    
    \end{tabular}}
    \caption{Summary of benchmarks for gender accuracy in machine translation. Dataset sizes listed are per-language (for benchmarks with different sizes per language, we take the smallest) and per-segment (sentence pair or document). 
    }
    \label{tab:small-comp}
\end{table*}

\section{Gender Evaluation Benchmarks for MT}
\label{sec:compare}

In this section, we review existing benchmarks on gender accuracy in machine translation, in order contextualize MT-GenEval with respect to similar work. 
Table~\ref{tab:small-comp} summarizes these benchmarks. 
Note that we focus our analysis on evaluation of gender translation accuracy in this section; see \citet{sun-etal-2019-mitigating} for a more general review of gender in natural language processing, and \citet{savoldi-etal-2021-gender} for a summary of work on gender bias in MT. 

\paragraph{WinoMT} One of the most widely-used datasets for evaluating gender accuracy in machine translation is WinoMT \citep{stanovsky-etal-2019-evaluating,kocmi-etal-2020-gender}, which was created by combining the Winogender \citep{rudinger-etal-2018-gender} and WinoBias \citep{zhao-etal-2018-gender} coreference test sets. 
As such, WinoMT contains synthetic Winograd-style sentences where gender is associated with pro-stereotypical and anti-stereotypical professions.\footnote{Recently, \citet{troles-schmid-2021-extending} extended WinoMT to include sentences with specific gender-stereotypical adjectives and verbs.} While the dataset does not contain reference translations, it instead includes a target language-specific alignment-based automatic metric. 
WinoMT is one of the gender translation accuracy benchmarks with the largest language coverage: the metric covers translation from English into AR, Czech (CS), DE, ES, FR, Hebrew (HE), IT, Polish (PL), RU, and Ukrainian (UK). 

\paragraph{MuST-SHE}
MuST-SHE~\citep{bentivogli-etal-2020-gender, savoldi-etal-2022-morphosyntactic} is a dataset of roughly 1,000 gender-specific segments for each of EN$\rightarrow$ES, FR, and IT.
Segments include both text and audio, and are extracted from TED talks~\citep{cattoni2021must}. 
The dataset contains segments where gender is disambiguated by the intra-sentential context, as well as segments where gender is only present as speaker metadata. It is curated to only include segments containing at least one gender-neutral source word that requires gender marking in the translation. The dataset also provides contrastive references for each segment. MuST-SHE was also extended with annotations of agreement chain and part of speech by \citet{savoldi-etal-2022-morphosyntactic}, and with source-side gender annotations by \citet{vanmassenhove-monti-2021-gender}.  

\paragraph{GeBioCorpus and Translated Wikipedia Biographies}
GeBioCorpus~\citep{costa-jussa-etal-2020-gebiotoolkit} and Translated Wikipedia Biographies\footnote{\url{https://ai.googleblog.com/2021/06/a-dataset-for-studying-gender-bias-in.html}} are closely related efforts that extract gender-related machine translation benchmarks from Wikipedia biographies. 
GeBioCorpus, which covers EN$\leftrightarrow$ES$\leftrightarrow$Catalan (CA), consists of gender-balanced parallel sentences that are automatically extracted and aligned using the GeBioToolkit.\footnote{\url{https://github.com/PLXIV/Gebiotoolkit}} 
A subset of 2,000 of the automatically extracted segments were reviewed by humans to yield an evaluation set. 
By contrast, Translated Wikipedia Biographies consists of professional human translations of Wikipedia biographies. 
It covers EN$\rightarrow$DE and EN$\rightarrow$ES and contains 138 documents, each with 8-15 sentences. 
This allows for evaluation of gender disambiguation in inter-sentential context. 
Note that both of these biography-based sets may contain irrelevant segments that have no gender information in either the source or the target. 

\paragraph{Other benchmarks}
\citet{renduchintala-etal-2021-gender} introduced SimpleGEN, which consists of relatively short, synthetic source phrases focusing on professions and verbs. The dataset is annotated for pro- and anti-stereotypicalness and probes for translations with ungrammatically mixed gender. 
Also based on professions is the occupations test set introduced by~\citet{escude-font-costa-jussa-2019-equalizing}, which consists of 1,000 human-translated EN$\rightarrow$ES sentence pairs that follow a simple pattern.

\section{Benchmarking Models with MT-GenEval}
\label{sec:benchmark}

In this section, we benchmark both commercial and research-scale models using MT-GenEval. 
In addition to giving initial baseline numbers for MT-GenEval, through these experiments we aim to show that MT-GenEval is a useful new benchmark. Specifically, we show that:
\begin{enumerate}
    \item MT-GenEval data is high-quality, and the benchmark is difficult even for state-of-the-art (SOTA) systems (section~\ref{context-benchmarking}).
    \item MT-GenEval measures a novel aspect of gender in translation that is absent from existing benchmarks (section~\ref{sec:benchmark-generic}).
    \item MT-GenEval is able to discriminate between models that are trained to improve contextual translation and translation of gender (section~\ref{sec:benchmarking-debias}).
\end{enumerate}

\subsection{Context-Level Gender Accuracy in Commercial Systems}
\label{context-benchmarking}
In this section, we evaluate three industrial systems on the contextual subset of the MT-GenEval benchmark. Our goal is to show that the dataset is sufficiently diverse and challenging even for systems trained on web-sized corpora. 

We used human evaluations to measure the gender accuracy of the translation outputs for each system for all eight target languages in the benchmark.
To anonymize the commercial systems, we label them \textsc{A}, \textsc{B}, and \textsc{C}. 
In order to anchor the evaluations, we additionally included reference translations in the evaluation.\footnote{Evaluators were not aware one of the translations was a human reference, and the order of the translations was shuffled.}

We asked two professional translators to judge gender accuracy in context\footnote{We showed translators both the context and the main sentence, but asked them to evaluate only the translation of the latter.} using a stratified sample of 600 source segments (100 for each sub-category in Table~\ref{tab:context-subsets}). We show the accuracy scores and inter-annotator agreement (IAA) computed with Krippendorff’s alpha~\citep{krippendorffalpha} in Table~\ref{tab:commer-genacct-contex}. We first note that IAA for the system outputs is high ($>$ 0.85) for the majority of target languages, with the exception of Hindi, Russian and French. For these three languages we also observe poor agreement ($\leq$ 0.25) on judging the correct reference translation. This indicates that some evaluators may have exhibited bias regarding gender accuracy judgments and over-penalized correct reference translations. For computing the system-level accuracy scores, we use the judgments of the ``preferred'' evaluator, which is the evaluator who judged the reference translation as correct more times than the other annotator. 

\begin{table}
    \centering
    \resizebox{\columnwidth}{!}{
    \begin{tabular}{l|c|c|c|c|c|c}
    & \multicolumn{4}{c|}{\textbf{Online Systems}} & \multicolumn{2}{c}{\textbf{Reference}} \\ \hline
        EN$\rightarrow$ & Acc-A & Acc-B & Acc-C & IAA & Acc & IAA \\ \hline
        AR &51.5 &51.3 & 50.5 & 0.93 & 95.8 & 0.72 \\
        FR &  56.1 & 55.9 & 56.2 & 0.82 & 98.3 & 0.25 \\
        DE &  52.3 & 51.3 & 54.0  & 0.96 & 92.2 & 0.84 \\
        HI &  59.7 & 61.1 & 61.3  & 0.70 & 97.1 & 0.25 \\
        IT &  55.9 & 54.9 & 55.7  & 0.86 & 97.4 & 0.65 \\
        PT &  50.9 & 52.3 & 52.7  & 0.89 & 91.4 & 0.68 \\
        RU &  56.9 & 57.2 & 57.7  & 0.77 & 97.0 & 0.16 \\
        ES &  57.0 & 58.2 & 58.5  & 0.94 & 97.6 & 0.89 \\ 
    \end{tabular}
    }
    \caption{Gender accuracy scores (\textit{Acc}) and inter-annotator agreement (\textit{IAA}) measured on the contextual subset for the three anonymized commercial systems and the correct reference translation.}
    \label{tab:commer-genacct-contex}
\end{table}

Across languages, the accuracy of these industrial systems is close to 50\%, ranging from 50.5\% for AR to a high of 61.3\% for HI. 
This indicates that they are largely not able to effectively take context into account to disambiguate the gender of the input. 
Thus, MT-GenEval is difficult even for SOTA industrial systems, despite the fact that they are trained on very large data that might actually include Wikipedia (which was used as the source for MT-GenEval).

Additionally, for the majority of languages, evaluators found the correct reference translations $>$ 95\% accurate, confirming the high quality of the dataset. However, for PT and DE, we observe a slightly lower reference accuracy (91.4\% and 92.2\%). 

\subsection{Evaluating Commercial Systems for Gender Quality Gap}
\label{sec:benchmark-generic}

Next, we evaluate the gender quality gap $\Delta_{qual}$ on the counterfactual test set for the same three industrial systems. 
These results are shown in Table~\ref{tab:commer-genqual}. 

\begin{table}
    \centering\small
    \begin{tabular}{l|r|r|r}
        EN$\rightarrow$ & \textbf{$\Delta_{qual}$ A} & \textbf{$\Delta_{qual}$ B} & \textbf{$\Delta_{qual}$ C} \\ \hline
        AR & 0.3 & 0.4 & 0.2 \\
        DE & 13.0 & 12.3 & 13.0 \\
        ES & 11.0 & 11.0 & 11.3 \\
        FR & 9.2 & 11.0 & 10.9 \\
        HI & 6.0 & 6.4 & 7.5 \\
        IT & 9.2 & 9.0 & 9.1 \\
        PT & 12.3 & 12.9 & 13.2 \\
        RU & 13.8 & 12.9 & 15.3 \\
    \end{tabular}
    \caption{Gender quality gap (lower magnitude is better) for the three anonymized commercial systems. Gender quality gap is defined as the difference in quality on the male and the female subsets (see section~\ref{sec:metrics-quality}).}
    \label{tab:commer-genqual}
\end{table}

With the exception of EN$\rightarrow$AR, the results show a clear pattern where the overall quality on masculine inputs is much higher (9.6 points on average) than the overall quality on feminine inputs (even though the inputs are identical aside from gender). 
Based on these results, as well as on initial observations regarding examples such as the one shown in Table~\ref{tab:qual-ex}, we ran a pilot analysis to see whether the automatically computed gender quality gap is indeed visible in generic (non-gender-related) quality as judged by humans. 
For EN$\rightarrow$DE, we extracted the 50 sentences from the test set that had the largest gap (i.e., $BLEU_{male} > BLEU_{female}$) and the smallest gap (i.e., $BLEU_{female} > BLEU_{male}$). 
A native German speaker manually checked whether the quality and gender translation accuracy differed between the female-referring and male-referring outputs. 
We found that most of the segments with a gender quality gap did indeed have meaningful differences in translation quality on portions of the segment unrelated to gender, even though those portions were (by design) identical in the source. 
Additionally, for the segments where male-referring translations were better, the gap in human-perceived quality was much larger than for the segments where female-referring translations were better. 

\begin{table}[htb]
    \centering
    \begin{tabular}{lp{6cm}}
        src: & We had to repair our relationship \textbf{because I wanted my \underline{mother}/\underline{father} back}. \\
        fem: & Wir mussten unsere Beziehung reparieren, \textbf{weil ich meine Mutter pflegte}.\\
        msc: & Wir mussten unsere Beziehung reparieren, \textbf{weil ich meinen Vater wollte}.\\
    \end{tabular}
    \caption{Model output from the \textsc{GFST-Ctx} system (section~\ref{sec:benchmarking-debias}) where the gender accuracy is correct, but the feminine (fem) input leads to a lower-quality output than the counterfactual masculine (msc) input. In the feminine translation, ``I wanted my mother back'' is translated incorrectly as ``I took care of my mother'', whereas the masculine translation is closer to the source: ``I wanted my father''.}
    \label{tab:qual-ex}
\end{table}

To our knowledge, the observation that there can be gaps in quality beyond gender-related words for otherwise equivalent inputs referring to different genders is novel.\footnote{A similar observation has been made for translation quality based on the demographics of the \textit{author} rather than the referred individual~\citep{hovy-etal-2020-sound,rabinovich-etal-2017-personalized,vanmassenhove-etal-2018-getting}, although such work could not enforce the inputs being otherwise equivalent.} 
Since MT-GenEval contains realistic and counterfactual data, it is now possible to evaluate models for these quality differences while controlling for content. 

\subsection{Contextual Gender Accuracy with Contextual and Gender-Balanced Models}
\label{sec:benchmarking-debias}

In this section, we use MT-GenEval to benchmark both contextual and gender-balanced NMT models trained on publicly available data. 
This helps us understand how existing methods for training these models affect performance on MT-GenEval. 
We focus on three language pairs: EN$\rightarrow$DE, EN$\rightarrow$FR, and EN$\rightarrow$RU. 
For each pair, we build four models:
\begin{itemize}
    \item \textsc{Base}: Non-contextual baseline 
    \item \textsc{Ctx}: Model trained with additional contextual data in the 2+2 format~\citep{tiedemann-scherrer-2017-neural}, following \citet{majumder2022multi}
    \item \textsc{GFST}: Model trained with additional gender-filtered self-training (GFST) data from \citet{choubey-etal-2021-gfst}\footnote{\url{https://github.com/amazon-research/gfst-nmt}}
    \item \textsc{GFST-Ctx}: Model trained with both the \textsc{GFST} data and the 2+2 \textsc{Ctx} data
\end{itemize}

The Transformer-base architecture \citep{vaswani2017attention} is used to train all the NMT models, with tied weight matrices for the source embeddings, target embeddings, and output layer. 
However, we use 8 decoder layers instead of 6, following the recommendation of \citet{majumder2022multi}. 
Training is done using Sockeye 3 \citep{hieber2022sockeye}. 
For training data, we use WMT19~\citep{barrault-etal-2019-findings} for EN$\rightarrow$DE and OpenSubtitles~\citep{lison-tiedemann-2016-opensubtitles2016} for EN$\rightarrow$FR and EN$\rightarrow$RU, all of which contain document-level data (used to train \textsc{Ctx} models). 
We use the dev sets from WMT19 (DE, RU) and IWSLT 2019 (FR) \citep{niehues-etal-2019-iwslt} for development. 

\begin{table}
    \centering\small
    \begin{tabular}{l|ccc|ccc}
    & \multicolumn{3}{c|}{\textbf{Contextual}} & \multicolumn{3}{c}{\textbf{Counterfactual}} \\ \hline
    EN$\rightarrow$ & DE & FR & RU & DE & FR & RU\\\hline
    \textsc{Base} & 66.7 & 66.1 & 62.5 & 71.0 & 63.0 & 79.7\\
    \textsc{Ctx} & 73.6 & \textbf{69.3} & 65.0 & 71.0 & 63.0 & 80.7\\
    \textsc{GFST} & 65.5 & 65.9 & 61.7 & 70.3 & 72.0 & 87.0\\
    \textsc{GFST-Ctx} & \textbf{77.1} & 68.8 & \textbf{68.1} & \textbf{76.0} & \textbf{75.3} & \textbf{91.0}\\
    \end{tabular}
    \caption{Automatic accuracy scores on MT-GenEval for the systems trained on public data.}
    \label{tab:public-res}
\end{table}
Table~\ref{tab:public-res} shows automatic accuracy scores for each system on both subsets of MT-GenEval (contextual and counterfactual).\footnote{Quality scores are shown in Appendix~\ref{sec:app-qual}.} 
On the contextual subset, as expected, we see much higher accuracy when a model is trained to take inter-sentential context into account: \textsc{Ctx} is better than \textsc{Base} and \textsc{GFST-Ctx} is better than \textsc{GFST}. This confirms that our test set is both sensitive to gender in context and challenging for vanilla contextual models (accuracy is below 75\%). We find that we can improve the accuracy of contextual models by combining gender-filtered and contextual data: \textsc{GFST-Ctx} is better than \textsc{Ctx} for German and Russian. On the other hand, gender balancing somewhat decreases the performance of non-contextual models (\textsc{Base} vs.\ \textsc{GFST}) on the contextual subset. 
Considering that the \textsc{Base} accuracy is higher than 50\%, this result indicates that non-contextual systems may be inferring gender from some source words that correlate with gender although they do not mark it explicitly (also observed for commercial systems in Table~\ref{tab:commer-genacct-contex}). 
Thus, the lower accuracy on the contextual set for \textsc{GFST} compared to \textsc{Base} could indicate a less gender-stereotypical model. 

On the counterfactual subset, the GFST data improves gender translation accuracy significantly overall, supporting the findings of \citet{choubey-etal-2021-gfst} on WinoMT and MuST-SHE. 
This confirms that our counterfactual subset is also sensitive to changes in gender balance in the training data. 
A surprising finding is that the contextual data improves the performance on non-contextual inputs in the counterfactual subset (\textsc{GFST-Ctx} is better than \textsc{GFST}). This aligns with prior work showing that adding contextual training data can introduce noise that acts as a regularizer~\citep{kim-etal-2019-document}, and that adding contextual data can help reduce gender bias in MT models~\citep{basta-etal-2020-towards}.

\section{Conclusions}
\label{sec:concl}
In this paper, we have introduced \textbf{MT-GenEval}, a counterfactual and contextual benchmark for evaluating gender accuracy in translation from EN into AR, DE, ES, FR, HI, IT, PT, and RU. 
In addition to the test data and evaluation metrics, we are releasing 2400 segments of development data.\footnote{We give suggested applications for the development set in Appendix~\ref{sec:gender8desc-uses}.} 
We have shown that this benchmark is useful for evaluating both commercial and research systems, including contextual machine translation models and gender-balanced models, in terms of gender accuracy as well as quality. 
We hope that this benchmark and development data will spur more research in the field of gender accuracy in translation on diverse languages.

\section*{Acknowledgments}
We would like to thank Tanya Badeka and Jen Wang for their help in creating the MT-GenEval data, Kathleen McKeown and Brian Thompson for their feedback on earlier drafts of this paper, and the anonymous reviewers for their suggestions.

\section{Limitations}
\label{sec:limit}

\paragraph{Dataset coverage}
The main limitation of this benchmark is that it only covers two human genders: female and male. 
Additionally, the language pairs covered in the benchmark all exhibit a similar pattern: language with limited grammatical gender (English) $\rightarrow$ language with morphological grammatical gender. 
It is not clear whether this benchmark could be expanded to more language pairs or used in the reverse direction. 
Finally, due to the counterfactual nature of the set, we excluded data containing individuals with different genders, as well as data with first names, which could bias the evaluation.

\paragraph{Annotator bias and errors}
In dataset creation, we relied heavily on human annotators, both for generating the counterfactual versions of the original sentences, and for translation into the target languages. 
Although we endeavored to mitigate biases in annotators by providing explicit instructions and examples, as well as by drawing annotators from a diverse population, it was not possible to eliminate such biases completely. 
For example, in the source annotation phase, an annotator created a male counterfactual of the sentence ``Pekgul trained as a nurse, a profession in which she worked both before and after her election as a politician.'' by changing ``nurse'' to ``male nurse''. 
This was prohibited by the instructions as the word ``nurse'' is already gender-neutral in English, and it exhibits the annotator's subtle bias that nurses are by default female. 
MT-GenEval could contain additional annotator errors, both in sources and in references. 

\paragraph{Limitations of the accuracy metric}
Our proposed accuracy metric relies on overlap with the reference, and as such it will not necessarily be reliable when translations of gendered words use synonyms that are not present in the reference. 
Additionally, this metric has yet not been compared to human scores for the counterfactual set due to unreliable human evaluations on that set.

\section{Ethical Considerations}
\label{sec:ethics}

In this paper, we release MT-GenEval, a benchmark for evaluating gender accuracy and quality in machine translation. 
We hope that this benchmark will be useful in evaluating representational harms related to gender in machine translation, particularly lower quality and accuracy in translation based on gender. 
The benchmark focuses on inputs that contain unambiguously gendered references to humans, and as such does not infer or assign gender in any way. 
Additionally, gender is not used as a variable in our work and we do not work with human subjects. 
We sourced our data from Wikipedia articles, which are publicly available and have a relatively low risk of inclusion of private information. 

As discussed in section~\ref{sec:limit}, the main limitation of our work is that evaluations are limited to two genders (female and male). 
We hope to be able to expand this work to more genders in the future. 

In creating our annotations, we worked with a language service provider to contract with professional translators and ensure suitable working conditions for them. 
Annotators were compensated in accordance with translation industry standards.

\bibliography{anthology,custom}
\bibliographystyle{acl_natbib}

\appendix
\section{Descriptive Statistics of MT-GenEval}
\label{sec:gender8desc-stats}

Table~\ref{tab:statistics-compare} shows representative data statistics comparing MT-GenEval (development subset) with WinoMT and MuST-SHE v1.0.
We select four languages for illustration. 
As can be observed, MT-GenEval is more diverse than the other two datasets, both when it comes to overall vocabulary as well as to the diversity of gendered phrases, particularly in the target langauges.

\section{Potential Uses for MT-GenEval Development Set}
\label{sec:gender8desc-uses}

In creating and releasing the MT-GenEval set, we hope to cover several potential use cases for improving and evaluating gender accuracy in translation. 
The counterfactual and contextual test sets allow gender translation accuracy to be evaluated on both the sentential and inter-sentential levels for translation from English into eight languages, with data releases for additional languages planned for the future. 
Additionally, translation evaluation in the reverse direction (i.e., *$\rightarrow$English) is possible for the counterfactual set, since this set was constructed so that most sentence pairs are marked for gender on both the source and the target sides. 

We anticipate that the 2,400 development sentences released for each language pair will be useful in training models to improve gender translation accuracy. 
Since the development set consists of gender-balanced counterfactual sentences, it can be used in gender fine-tuning as introduced by~\citet{saunders-byrne-2020-reducing}, with the added advantage that the MT-GenEval development data is naturally occurring and more complex than artificially constructed segments used in their original work. 
As an alternative, this data can be potentially used to train a model that generates counterfactuals automatically, instead of relying on rule-based gender counterfactuals as in prior work. 
Other prior work on improving gender in translation used wordlists and morphological taggers to extract gender-specific data from a generic corpus~\citep{choubey-etal-2021-gfst}; the counterfactual MT-GenEval development data that we are releasing could generalize this process by being used to train a classifier that automatically detects gender-specific segments.

\section{Quality Scores on Contextual and Gender-Balanced Models}
\label{sec:app-qual}

Table~\ref{tab:public-qual} shows gender-specific BLEU scores on the models trained on public data from section~\ref{sec:benchmarking-debias}. 
Unlike for the commercial systems (section~\ref{sec:benchmark-generic}), we do not see a large gender quality gap in these models. 
However, BLEU scores are quite low, particularly for EN$\rightarrow$RU, possibly due to domain mismatch (Wikipedia vs.\ WMT/OpenSubtitles). 

\begin{table}[htb]
    \centering\small
    \begin{tabular}{l|ccc|ccc}
    & \multicolumn{3}{c|}{$BLEU_{female}$} & \multicolumn{3}{c}{$BLEU_{male}$} \\ \hline
    EN$\rightarrow$ & DE & FR & RU & DE & FR & RU\\\hline
    \textsc{Base} & {12.9} & 12.9 & 8.0 & 12.7 & 13.4 & 8.7\\
    \textsc{Ctx} & 11.1 & 12.8 & 6.8 & 11.7 & 13.8 & 7.8\\
    \textsc{GFST} & 12.7 & 13.8 & {8.5} & {13.6} & 14.4 & {9.5}\\
    \textsc{GFST-Ctx} & 12.8 & {14.5} & 7.6 & 13.2 & {15.2} & 8.6\\
    \end{tabular}
    \caption{Automatic accuracy scores on the contextual and counterfactual subsets for the systems trained on public data.}
    \label{tab:public-qual}
\end{table}

\begin{table*}[th]
    \centering
    \begin{tabular}{l|l|r|r|r|r|r}
        \multicolumn{3}{c}{} & \multicolumn{2}{|c}{\textbf{Source}} & \multicolumn{2}{|c}{\textbf{Target}} \\
        \hline
        Dataset & EN & \# instances & \# distinct & \# distinct gendered  & \# distinct & \# distinct gendered \\
        & $\rightarrow$ & & \multicolumn{1}{c|}{words} & \multicolumn{1}{c|}{phrases} & \multicolumn{1}{c|}{words} & \multicolumn{1}{c}{phrases} \\
        \hline
        & AR & & 6,690 & 350 & 11,194 & 3,890 \\
        MT-GenEval & DE & & 6,604 & 328 & 8,053 & 1,091 \\
         & FR & 1,200 & 6,575 & 369 & 7,640 & 1,480 \\
        & RU & & 6,619 & 348 & 10,121 & 2,253 \\
        \hline
        WinoMT & -- & 1,944 & 1,883 & -- & -- & -- \\
        \hline
        MuST-SHE & FR & 1,113 & 4,605 & -- & 5,792 & 1,456 \\
    \end{tabular}
    \caption{Representative data statistics of MT-GenEval, WinoMT, and MuST-SHE v1.0.}
    \label{tab:statistics-compare}
\end{table*}

\end{document}